\documentclass{article}

\usepackage{times}
\usepackage{graphicx} 
\usepackage{subfigure} 

\usepackage{natbib}

\usepackage{algorithm}
\usepackage{algorithmic}

\usepackage{hyperref}
\usepackage{acronym}
\acrodef{DL}{Deep Learning}
\acrodef{RNN}{Recurrent Neural Network}
\acrodef{RPS}{Renewable Portfolio Standard}
\acrodef{CNN}{Convolutional Neural Network}
\acrodef{GPU}{Graphics Processing Unit}
\acrodef{SGD}{Stochastic Gradient Descent}
\acrodef{LSTM}{Long Short Term Memory}
\acrodef{BP}{Back Propagation}
\acrodef{NLP}{Natural Language Processing}
\acrodef{DL-STF}{DL-based Spatio-Temporal Forecasting}

\acrodef{CoM}{Concentration of Measure}
\acrodef{i.i.d.}{independent and identically distributed}
\acrodef{LTI}{Linear Time-Invariant}
\acrodef{LTV}{Linear Time-Variant}
\acrodef{LPV}{Linear Parameter-Varying}
\acrodef{RIP}{Restricted Isometry Property}
\acrodef{SVD}{Singular Value Decomposition}
\acrodef{CS}{Compressive Sensing}
\acrodef{DSP}{Digital Signal Processing}
\acrodef{CSI}{Compressive System Identification}
\acrodef{CTI}{Compressive Topology Identification}
\acrodef{CBD}{Compressive Binary Detection}
\acrodef{OMP}{Orthogonal Matching Pursuit}
\acrodef{MP}{Matching Pursuit}
\acrodef{ERC}{Exact Recovery Condition}
\acrodef{BOMP}{Block Orthogonal Matching Pursuit}
\acrodef{COMP}{Clustered Orthogonal Matching Pursuit}
\acrodef{CoSaMP}{Compressive Sampling Matching Pursuit}
\acrodef{KKT}{Karush-Kuhn-Tucker}

\acrodef{FIR}{Finite Impulse Response}
\acrodef{DFT}{Discrete Fourier Transform}
\acrodef{DCT}{Discrete Cosine Transform}
\acrodef{JL}{Johnson-Lindenstrauss}
\acrodef{ROC}{Receiver Operating Curve}
\acrodef{NP}{Neyman-Pearson}
\acrodef{ARX}{Auto Regressive with eXternal input} 
\acrodef{MISO}{Multi-Input Single-Output}
\acrodef{SISO}{Single-Input Single-Output}
\acrodef{MIMO}{Multi-Input Multi-Output}

\acrodef{LASSO}{Least Absolute Shrinkage and Selection Operator}
\acrodef{GLASSO}{Group LASSO}
\acrodef{NNG}{Non-Negative Garrote}
\acrodef{LARS}{Least Angle Regression}
\acrodef{I/O}{Input/Output}
\acrodef{CST-WSF}{Compressive Spatio-Temporal Wind Speed Forecasting}
\acrodef{AR}{Autoregressive}
\acrodef{M-AR}{Multivariate Autoregressive}
\acrodef{NM-AR}{Nonuniform Multivariate Autoregressive}

\acrodef{NWP}{Numerical Weather Prediction}
\acrodef{TDD}{Trigonometric Direction Diurnal}
\acrodef{RSTD}{Regime Switching Space-Time Diurnal}
\acrodef{TDDGW}{TDD with Geostrophic Wind Information}
\acrodef{ACK}{Nantucket Memorial Airport}
\acrodef{WT}{Wavelet Transform}
\acrodef{ANN}{Artificial Neural Network}
\acrodef{MAE}{Mean Absolute Error}
\acrodef{RMSE}{Root Mean Squared Error}
\acrodef{NRMSE}{Normalized Root Mean Squared Error}
\acrodef{TSOs}{Transmission System Operators}
\acrodef{IPPs}{Independent Power Producers}
\acrodef{ST-ANN}{Spatio-Temporal ANN}
\acrodef{LS}{Least Squares}
\acrodef{METAR}{Meteorological Terminal Aviation Routine}

\usepackage[utf8]{inputenc} 
\usepackage[T1]{fontenc}    
\usepackage{url}            
\usepackage{booktabs}       
\usepackage{amsfonts}       
\usepackage{nicefrac}       
\usepackage{microtype}      
\usepackage{multirow}

\usepackage{graphicx}
\graphicspath{ {imgs/} }
\usepackage{amsmath}
\usepackage{amssymb}
\usepackage{pgf}
\usepackage{color}

\newcommand{\revised}[1]{{\color{black} #1}} 
\newcommand{\cut}[1]{}

\usepackage[accepted]{icml2017}

\icmltitlerunning{Deep Forecast:Deep Learning-based Spatio-Temporal Forecasting}

\begin{document} 

\twocolumn[
\icmltitle{Deep Forecast: \\
Deep Learning-based Spatio-Temporal Forecasting}





\begin{icmlauthorlist}
\icmlauthor{Amir Ghaderi}{uta}
\icmlauthor{Borhan M. Sanandaji}{cu}
\icmlauthor{Faezeh Ghaderi}{uta}
\end{icmlauthorlist}

\vskip 0.2in
\hskip 2in 
\it{To the memory of Maryam Mirzakhani (1977-2017)}

\icmlaffiliation{uta}{University of Texas at Arlington, Texas, USA}
\icmlaffiliation{cu}{Mojio Inc., Palo Alto, CA, USA}

\icmlcorrespondingauthor{Amir Ghaderi}{amir.ghaderi@mavs.uta.edu}


\vskip 0.3in
]



\printAffiliationsAndNotice{} 

\begin{abstract} 
The paper presents a \emph{spatio-temporal} wind speed forecasting algorithm using \ac{DL} and in particular,  \acp{RNN}. Motivated by recent advances in renewable energy integration and smart grids, we apply our proposed algorithm for wind speed forecasting. Renewable energy resources (wind and solar) are random in nature and, thus, their integration is facilitated with accurate short-term forecasts. 
In our proposed framework, we model the spatio-temporal information by a graph whose nodes are data generating entities and its edges basically model how these nodes are interacting with each other. One of the main contributions of our work is the fact that we obtain forecasts of all nodes of the graph at the same time based on one framework. 
Results of a case study on recorded time series data from a collection of wind mills in the north-east of the U.S. show that the proposed \ac{DL}-based forecasting algorithm significantly improves the short-term forecasts compared to a set of widely-used benchmarks models.

\end{abstract} 

\acresetall
\section{Introduction}

\subsection{Variable Energy Resources}
\revised{Many countries in the world and many states in the U.S. have mandated aggressive \acp{RPS}. Among different renewable energy resources, wind energy itself is expected to grow to provide between $15$ to $25\%$ of the world's global electricity by 2050. According to another study, the world total wind power capacity has doubled every three years since 2000, reaching an installed capacity of 197 GW in 2010 and 369 GW in 2014~\cite{CA_renewable_portfolio}, \cite{iea2013technology}.
The random nature of wind, however, makes it difficult to achieve the power balance needed for its grid integration~\cite{smith2007utility}.
The use of ancillary services such as frequency regulation and load following  to compensate for such imbalances is facilitated by accurate forecasts~ \cite{hao2013aggregate}, \cite{sanandaji2014improved}.}


\subsection{Main Contributions} 
\revised{We present a \emph{spatio-temporal} wind speed forecasting algorithm using \ac{DL} and in particular, ~\acp{RNN}. 
In our proposed framework, we model the spatio-temporal information by a graph whose nodes are data generating entities and its edges model how these nodes are interacting with each other. One of the main contributions of our work is the fact that we obtain forecasts of all nodes of the graph at the same time and using one framework. One of the most important points is that we do not know the relationship between stations and the trained model determines which stations are more important to forecast one specific station. Our code and data are available at \href{https://github.com/amirstar/Deep-Forecast}{https://github.com/amirstar/Deep-Forecast}. 
}
\cut{ We summarize the contributions of the paper as:
\begin{enumerate}
\item Available methods in this area can forecast wind speed only for one station but our framework can forecast all output for all stations and show awesome 
results just with one frame work learned on data. Our model uses all information in the graph to forecast output. 
\item One of the most important points is that we do not know the relationship between locations of the stations. The model learned spatial relation between stations.
\item We define forecasting as a new simple and effective framework and watch at it from new point of view. We introduce graph forecasting instead of node
forecasting.
\item The model is suitable not only for wind forecasting but also for many other applications especially in control systems and forecasting e.g. traffic prediction ... .
\item We perform extensive experiments on the data to show the performance of the algorithm in different situations.
\end{enumerate}
In what follows, we explain our proposed \ac{DL}-based framework in details.}
\subsection{Wind Energy Forecasting Methods} 
\revised{One can directly attempt to forecast wind power.~An alternative approach is to forecast the wind speed and then convert it to wind power using given power curves. This approach will accommodate different wind turbines installed in a wind farm  experiencing the same wind speed profile but resulting in different wind power generation. We focus on wind speed forecasting in this paper. 
Wind speed forecasting methods can be categorized to different groups: (i) model-based methods such as \ac{NWP} vs. data-driven methods, (ii) point forecasting vs. probabilistic forecasting, and (iii) short-term forecasting vs. long-term forecasting.  This paper is concerned with short-term point forecasting using both temporal data as well as spatial information. For a more complete survey of wind speed forecasting methods see~\cite{zhu2012short} and~\cite{tascikaraoglu2014review}, among others.}


\section{Related works}
%
\subsection{Spatio-Temporal Wind Speed Forecasting}

There is a growing interest in the so-called \emph{spatio-temporal} forecasting methods that use information from neighboring stations to improve the forecasts  of a target station, since there is a significant cross-correlation between the time series data of a target station and its surrounding stations. We review some of the spatio-temporal forecasting methods.
\cite{gneiting2006calibrated} introduced the \ac{RSTD} model for average wind speed data based on both spatial and temporal information. This method was later improved by Hering and Genton~\cite{hering2010powering} who incorporated  wind direction in the forecasting process by introducing \ac{TDD} model. 
\cite{xieshort} also considered probabilistic \ac{TDD} forecast for power system economic dispatch. 
\cite{dowell2013short} employed a multi-channel adaptive filter to predict the wind speed and direction by taking advantages of spatial correlations at numerous geographical sites.
\cite{he2014spatio} presented Markov chain-based stochastic models for predictions of wind power generation after characterizing the statistical distribution of aggregate power with a graph learning-based spatio-temporal analysis.
Regime-switching models based on wind direction are studied by~\cite{tastu2011spatio} where they consider various statistical models, such as ARX models, to understand the effects of different variables on forecast error  characteristics.
A methodology with probabilistic wind power forecasts in the form of predictive densities taking the spatial information into account was developed in~\cite{tastu2014probabilistic}. 
Sparse Gaussian Conditional Random Fields (CRFs) have also been deployed for probabilistic wind power forecasting~\cite{wytock2013largescale}.
See~\cite{zhang2014review} for a comprehensive review of the state-of-the-art methods.

\subsection{Forecasting using Neural Networks}
Among different \ac{DL} algorithms, \ac{RNN} has been commonly used in forecasting applications. \cite{gravestext} used \ac{LSTM} in text generation and predict one output in each time step. He used input values till time $t$ to get prediction
in time $t+1$.
\cite{wang2016deep}) used a mixture of wavelet transform, deep belief network, and spine quantile regression for wind speed forecasting. 
\cite{grover2015deep} proposed a hybrid model using deep neural network. \cite{ma2017generalized}) used fuzzy logic and neural networks for forecasting. \cite{li2010comparing} proposed a comparison on three neural networks for 1-hour wind speed forecasting.
\\

There are two important differences between our proposed method compared to other methods: 1) existing methods forecast output of one node while our approach yields in forecasts of all nodes and, 2) most of the existing methods update during the input horizon and use the new data but our model 
does not need to update during input horizon which can improve the speed and performance of the algorithm.

\section{Recurrent Neural Networks and LSTM}
%
\revised{Originating from computer vision and image classification, \ac{DL} has shown promising results in different tasks 
in recent years~(\cite{hintonImagenet}, \cite{faster}, \cite{sequence}).~Its ability in handling large amount of data and learning nonlinear and complicated models has made it an appealing framework. 
In one of the earliest works,~\cite{hintonImagenet} proposed to run a deep (a neural network with several hidden
layers) \ac{CNN} on a \ac{GPU} to classify a large data set of images (ImageNet dataset,~\cite{imagenet}). 
Among several algorithms that have been proposed in \ac{DL} for different tasks, \ac{RNN} is proposed for modeling temporal data and has been applied to speech recognition, activity recognition, \ac{NLP}, etc. In the following, we provide some insights on how an \ac{RNN} is built.}

\revised{Let $\mathcal{X} \triangleq \{x_1, x_2, \dots, x_\ell\}$ be a sequence of data where $x_t$ is the vector of features at time $t$ and $\ell$ is the input horizon. There exist many variations for the \ac{RNN} structure.  Some structures generate output for each time step while there are \acp{RNN} with one final output at time $\ell$ when $\mathcal{X}$ is applied as an input to the \ac{RNN}.
Let $\mathcal{Y} \triangleq \{y_1, y_2, \dots, y_\ell\}$ be the sequence of outputs at each time step.}~\revised{A function $f$ is applied on each input $x$ and the output of $f$ in the previous time step. One should note that the same function should be used during all time steps. This is an important point which makes the model capture the useful information content of the data (used for training) at each time step.
\ac{SGD} and \ac{BP} are used to train the function and find optimal parameters.}
\\

\begin{figure}[h]
\centering
\includegraphics[scale=0.45]{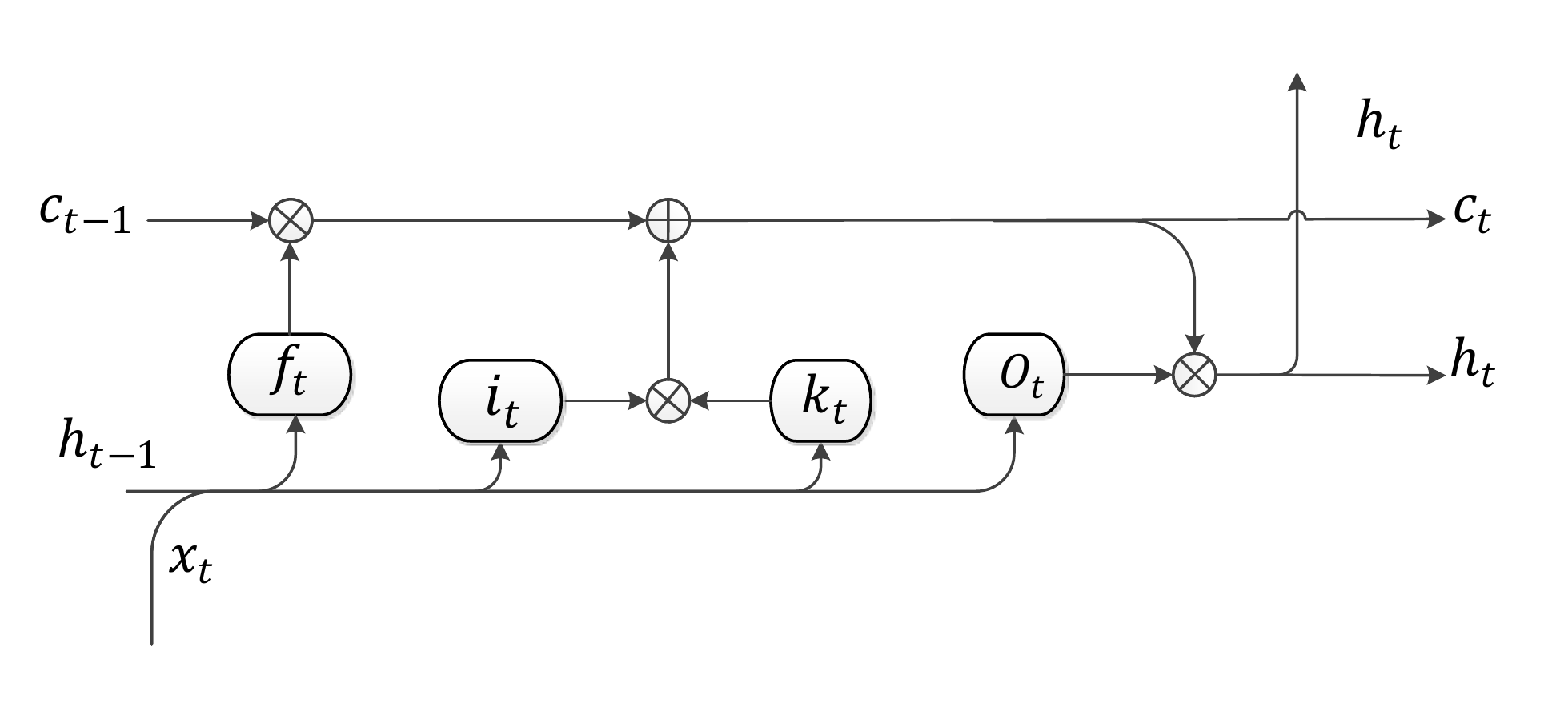}
\caption{LSTM block at time $t$ }
\label{fig:LSTMBlock}
\end{figure}

\revised{There exist some issues with the basic \ac{RNN} structure such as vanishing gradient (especially for long input sequences).~\cite{lstm} proposed \ac{LSTM} 
to address such problems. In short,~\ac{LSTM} provides a framework to embed the required information for the function.~\ac{LSTM} networks have better convergence performance 
compared to the basic \ac{RNN}.
\ac{LSTM} consists of multiple functions as compared to one function in vanilla \ac{RNN}. These functions
try to remember the helpful and forget the unnecessary information from inputs. Figure \ref{fig:LSTMBlock} shows relationship between functions in \ac{LSTM}.
The output of each step is calculated following the formulas provided in~(\ref{eq:lstm}): 
\begin{equation}
\label{eq:lstm}
\begin{split}
&f_t = g(W_f.x_t + U_f.h_{t-1} + b_f)
\\
&i_t = g(W_i.x_t + U_i.h_{t-1} + b_i)
\\
&k_t = tanh(W_k.x_t + U_k.h_{t-1} + b_k)
\\
&c_t = f_t \times c_{t-1} + i_t \times k_t
\\
&o_t = g(W_o.x_t + U_o.h_{t-1} + b_o)
\\
&h_t = o_t \times tanh(c_t)
\end{split}
\end{equation}

Where $x_t$ is the input vector at time $t$ and $g$ is an activation function like $Sigmoid$ or $ReLU$. $W$, $U$ are weight matrices and $b$ is the bias vector. $h_t$ and $c_t$ are output and cell state vector at time $t$. $f_t$ has served for remembering old information and $i_t$ has served for getting new information.There are many variations of \ac{LSTM}. Keen readers can find more about \ac{LSTM} in \cite{goodfellow2016deep}.}

%
\section{DL-based Spatio-Temporal Forecasting (DL-STF)}
\revised{In this section, we outline our proposed spatio-temporal forecasting scheme which is based on \ac{DL}. We namely call our algorithm \ac{DL-STF}. Let graph $G$ be defined as $G \triangleq [E,V]$ where $E$ denotes the edges and $V = \{v_1, \dots, v_n\}$ denotes the nodes. Each node of the graph $v_i$ generates data at each time step. Node $v_i$ at time $t$ generates $x^t_i$ which is a scalar (e.g., wind speed in our problem).}
Assume $x^t_i$ is sampled from an unknown distribution $P_{real}(x)$. Also $s^t = [x^t_1, x^t_2\dots x^t_n]$ (it is a vector) contains the output of all nodes at time $t$.Similarly,  let $\hat{x}^t_i$ be the output prediction of node $i$ at time $t$. 
\cut{$\hat{x}^t_i$ has the same dimension as $x^t_i$ and is supposed to come from an unknown distribution 
$P_{model}(x)$. This assumption seems questionable to me} Let $\hat{s}^t = [\hat{x}^t_1, \hat{x}^t_2\dots \hat{x}^t_n]$ be the vector containing prediction of all nodes (same size as $s^t$).
We assume we only have real data for all nodes every $h$ time steps.
Our goal is to 
predict $s^{t}$, using \{$s^{k}\} , k \in \{t-\ell,t-\ell+1,... ,t-1\}$. Based on moving horizon scheme, when real value for $s^{k}$ is not available, we use its prediction, $\hat{s}^k$.
 
\revised{
\subsection{Time step models} 
We have access to real data every $h$ hours and want to forecast wind speed for the next $h$ hours. In different time steps we have different kind of inputs. For the first time step, we have real data for all inputs but for the next time step, we have real data for all inputs except one. For that one we use forecast data from previous step. This scheme repeats for all the $h$ steps. Based on this paradigm, we define a specific model for each time step over the input horizon. In order to train the model, we use real values as much as we can, but if we do not have real values, we use forecast values from previous trained models. So for example the model for forecasting at time $t + 2$ should differ from model for forecasting at time $t + 3$. 
If we have $n$ stations, the input for our algorithm is an $n$-dimensional vector at each time step and we forecast an $n$-dimensional vector for
next time step. 

Let $t$ denote the global time index, $h$ the number of time steps in moving horizon, and $\ell$ the input horizon. We train $h$ different models. Model number $i$ is represented by $M_i$. We define $\hat{t} = t ~ \text{mod}~ h$ and the relation between $\hat{t}$ and $i$ is as follows: 

\begin{equation}
\label{eq:i}
i = \begin{cases}\hat{t}  & ,\text{if} ~ \hat{t} \neq 0 \\h & ,\text{if} ~ \hat{t} = 0 \end{cases}
\end{equation}

The input and output of each model is different from others. Also structure, number of parameters, and details of each model could be 
different from others. This flexibility in defining and customizing the models based on the time step during the prediction horizon ($h$) is one of the strength of our framework. For $i=1, 2, \cdots, h$, we have: 
\begin{itemize}
\item Model:  $M_i$
\item Input: $\{s^{t-\ell}, s^{t-\ell+1}, ...,~ s^{t-i}, \hat{s}^{t-i+1}, ..., \hat{s}^{t-1} \}$
     
($\ell-i+1$ real values, $i -1$ forecasted  values)
\item Algorithm: \ac{RNN} with LSTM blocks
\item Output: Forecast of output of all nodes, $\hat{s}^{t}$
\end{itemize}
$\hat{s}^{t}$ is output of the model $M_i$, which $i$ is calculated from (\ref{eq:i}).
If $t-l$ is greater than $t-i$ then we only use the last $l$ predicted  values. Also if $t-i+1$ is equal to $t$, only real values are used. Loss 
function for each model $M_i$ is defined as $L(M_i(\{s^{t-\ell},... ~ s^{t-i}, \hat{s}^{t-i+1},... ~ \hat{s}^{t-1} \}), x^t)$ where $x^t$ is the real 
output of the nodes in the graph at time $t$ and $L(a,b)$ is the mean absolute error of $a$ and $b$. In figure \ref{fig:model} we  show  the  overview  of the model's detail.

\begin{figure}[h]
\centering
\includegraphics[scale=0.25]{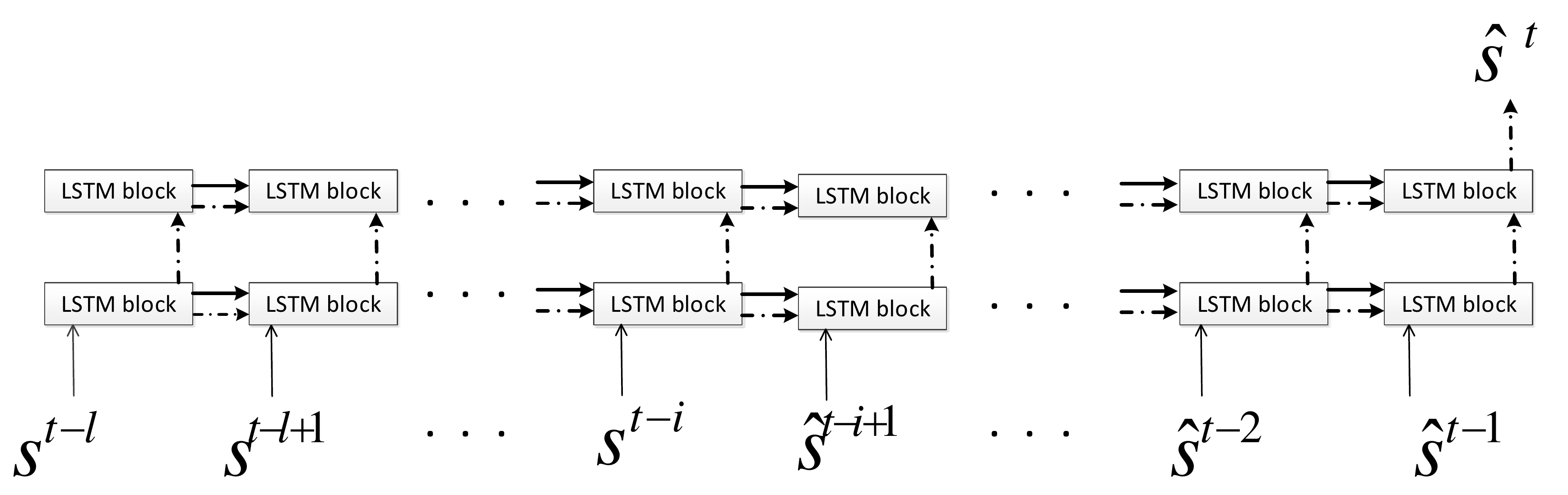}
\caption{The model $M_i$ trained at time t. Inputs are $s^{t-\ell}, s^{t-\ell+1}, ...,~ s^{t-i}$ (real values) and  $\hat{s}^{t-i+1},... ~ \hat{s}^{t-1}$ (forecasted values from previous trained models). 
Black thick and dashed arrows are $c_t$ and $h_t$ based on Figure \ref{fig:LSTMBlock}. } 

\label{fig:model}
\end{figure}

}

\section{Case Study of 57 Stations in East Coast}
\label{sec:case}
We apply \ac{DL-STF} to real wind speed data. East coast states are good candidates for our study as: (i) wind speed profiles are higher and (ii) there are more stations in a close vicinity in these states.  
\subsection{Data Description}
\begin{figure}[tb]
\begin{center}
\includegraphics[width =1\columnwidth]{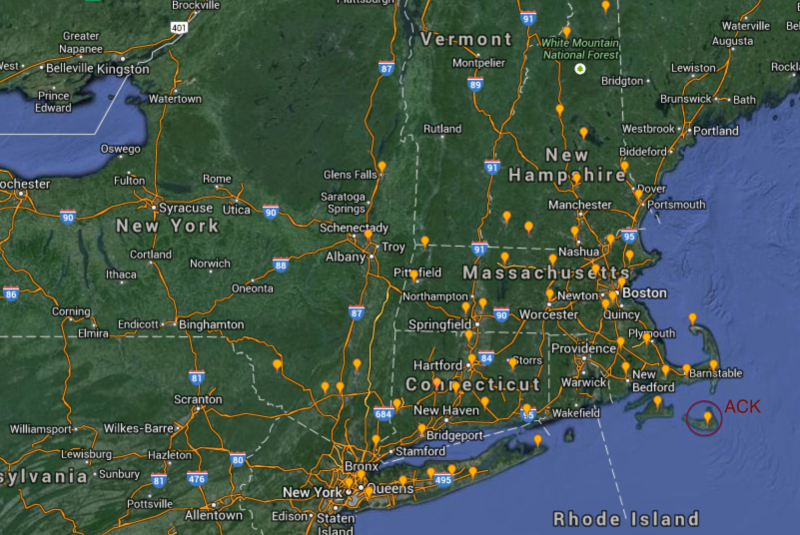}
\end{center}
\caption{Map of the area under study. The $57$ measuring locations in east coast are shown with yellow points. Circled in red is the target station ACK.}
\label{fig:stations_map}
\end{figure}

We use hourly wind speed data from \ac{METAR} weather reports of 57 stations in east coast including Massachusetts, Connecticut, New York, and New Hampshire~\cite{Iowa}. 
Fig.~\ref{fig:stations_map} depicts the area under study and the location of these 57 stations. 
The target station~\ac{ACK} (circled in red) is located on an island and is subject to wind profiles with high ramps and speeds due to the fact that the surrounding surface has very low roughness heights. Furthermore, this area has good correlations with other stations owing to the fact the prevailing wind direction of this region is mainly northwest or southeast. 
A time period from January 6, 2014 to February 20, 2014 is considered as test set in our simulations. This time period has the most unsteady wind conditions throughout the year. 
\\

\subsection{Results and Discussion}

\begin{figure}
 \centering
 \includegraphics[scale=0.45]{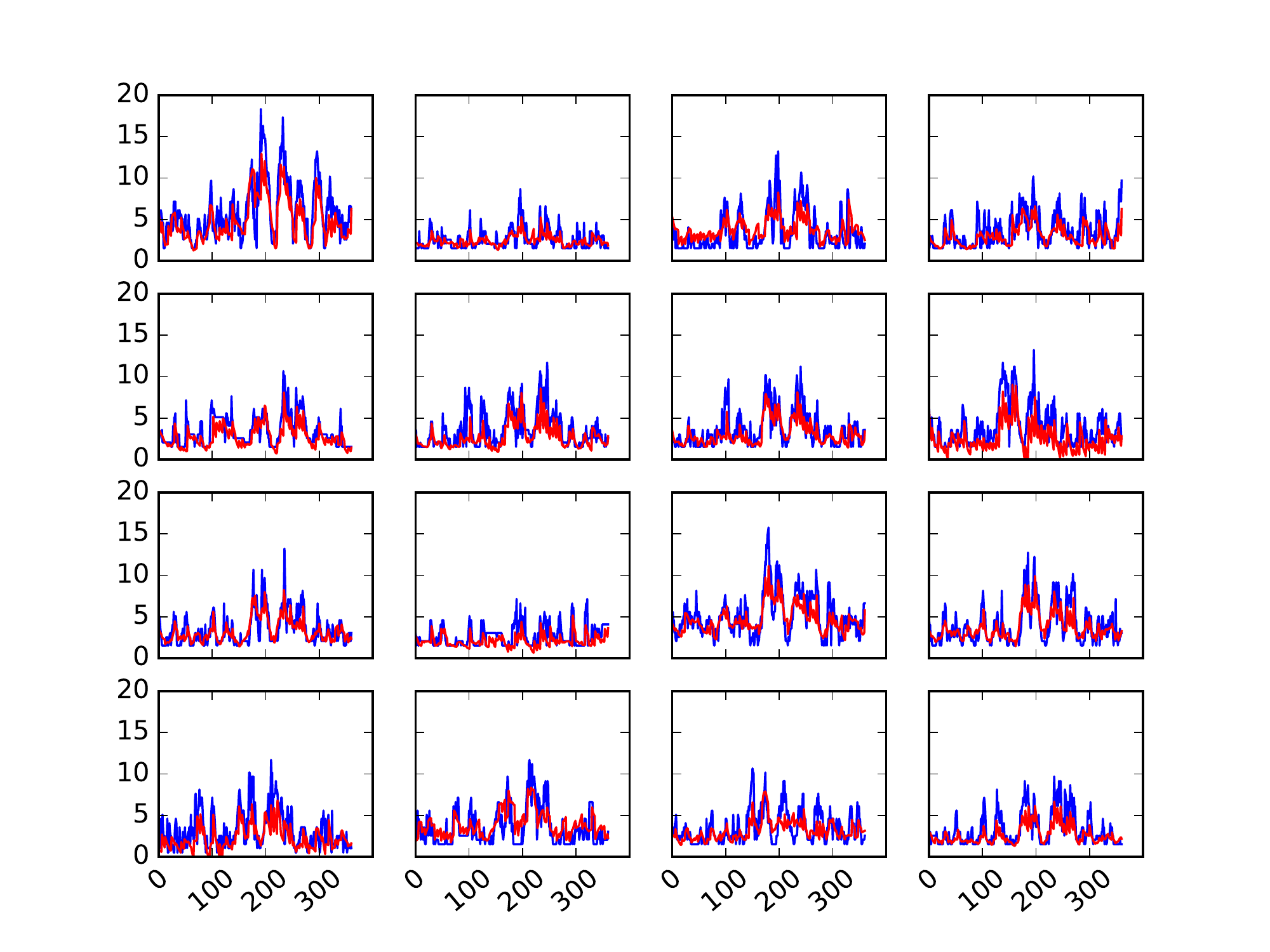} 
 \caption{Forecasting performance on 16 stations. Blue lines are real values and red ones are forecast values. Horizontal axis shows time steps and vertical axis shows wind speed(m/s).}
 \label{fig:finalEpoch}
\end{figure}

\revised{In this section we discuss the details about our implementation and hyper-parameters setting. In our experiments $h = 6$ and we chose 
$\ell = 12$ based on a cross validation study. The optimizer is MSRProp which shows good performance for \ac{RNN} with learning rate of $0.001$. The activation function is ReLu. Data is normalized between 0 and 1. We use one fully connected layer on top of $h_t$ features to create the desired output layer.

We use TensorFlow \cite{tensorflow2015-whitepaper} and Keras \cite{chollet2015keras} and our code and data are available online. 
For models whose input includes predicted values, we need to increase the model capacity to help overcome over-fitting. Thus, we increase the number of neurons and layers and use stacked \ac{LSTM}.}
\revised{To show performance of our algorithm we use three common error measure MAE, RMSE, and NRMSE. 
Table \ref{tab:errorsOneNode} shows comparison of three common error measures between proposed method and other methods. It is worth mentioning that other existing methods are trained to forecast only one node at a time but our method can forecast the output of all nodes in the graph at one time.
More importantly, as we can see the error comparisons, our method has smaller error values compared to all other methods and outperform state-of-the-art results. More details about how other methods work are available in \cite{sanandaji2015low}, \cite{tascikaraoglu2016exploiting}.

\begin{table}[!h]
\centering
\caption{Error measures for different methods for one node (ACK)}
\begin{tabular}{|c|c|c|c|} \hline
\multirow{2}{*}{Method} & MAE    & RMSE  & NRMSE  \\ 
       & (m/s)      & (m/s)     & (\%) \\ \hline
 Persistence Forecasting    &    2.14  & 2.83  & 16.86 \\ \hline
 AR of order 1& 2.07    &     2.76 &16.44 \\ \hline
AR of order 3 & 2.07  & 2.76 & 16.40 \\ \hline
WT-ANN & 1.82  & 2.47 & 14.68 \\ \hline
AN-based ST & 1.80  & 2.30  & 13.69 \\ \hline
LS-based ST & 1.72  & 2.20 & 13.08 \\ \hline
DL-STF &  1.63      &    2.19         &      \textbf{13.08}      \\ \hline
DL-STF(All nodes) &  \textbf{1.18}      &    \textbf{1.62}         &      16.28      \\ \hline
\end{tabular}
\label{tab:errorsOneNode}
\end{table}

Table \ref{tab:errorsAllNodes} shows the average of three error measures for all nodes in the graph. To the best of our knowledge there is no other method capable of forecasting outputs of all nodes in a graph in one framework. The average of the error measures of all nodes is even better than error measures for one node(ACK) with relative improvement about 27\% for MAE and RMSE. 

\begin{table}[h!]
  \caption{Average error measures over all locations using \ac{DL-STF}}
  \label{tab:errorsAllNodes}
  \centering
  \begin{tabular}{cccc}
    \midrule 
          & MAE(m/s) & RMSE(m/s) & NRMSE(\%) \\
    \midrule
	Our method & 1.18  & 1.62 & 16.28     \\
    \bottomrule
  \end{tabular}
\end{table}

In \ac{DL-STF}, we model all information and the hidden interactions between nodes of the graph. As explained earlier, in a spatio-temporal setting we use information of all nodes to forecast one node's
output in order to improve the forecasting performance as compared to the case when we only use one node's data (temporal setting). Table \ref{tab:oneStationVSAllStations} illustrates a comparison
between these two cases: 1) \ac{DL-STF} trained on all nodes of the graph to forecast one node's output (node ACK) and 2) \ac{DL-STF}  trained with data only from node ACK and, thus, we don't count for hidden relationships between nodes. Table \ref{tab:oneStationVSAllStations} shows that the error measures in case 1 (spatio-temporal forecasting) has significantly improved. 
\\

\begin{table}[h]
  \caption{Comparison of forecasting error measures. \textbf{First row}: train only on data from the node ACK and test on the node ACK. \textbf{Second row}: train on data from all nodes of the graph and test on the node ACK. \textbf{Third row}: train on data from all nodes and test on all nodes and calculate the average error measures over all nodes.}
  \label{tab:oneStationVSAllStations}
  \centering
  \begin{small}
  \begin{tabular}{cccc}
    \midrule 
          & MAE(m/s) & RMSE(m/s) & NRMSE(\%) \\
    \midrule
    one node & 1.99  & 2.60 & 15.46  \\
    
    \midrule
    all nodes (ACK) & 1.63  & 2.19 & \textbf{13.08}  \\

    \midrule
    mean all nodes & \textbf{1.18}  & \textbf{1.62} & 16.28   \\
    
    \bottomrule
  \end{tabular}
\end{small}	
\end{table}

In Figure \ref{fig:finalEpoch} we show the accuracy of forecasting for 16 nodes of the graph. It shows 
real values vs forecast values on test data. The average of error measures are 1.203, 1.663, 16.378 for MAE, RMSE, NRMSE respectively.}
\\

\section{Conclusion}
\revised{A \emph{spatio-temporal} wind speed forecasting algorithm using \ac{DL} is presented. 
In our proposed framework, we model the spatio-temporal information by a graph whose nodes are data generating entities and its edges basically model how these nodes are interacting with each other. One of the main contributions of our work is the fact that we obtain forecasts of all nodes of the graph at the same time. 
Results of a case study on recorded time series data from a collection of wind mills in the north-east of the U.S. show that the proposed \ac{DL}-based forecasting algorithm significantly improves the short-term forecasts compared to a set of widely-used benchmarks models.} 


\bibliography{example_paper}
\bibliographystyle{icml2017}

\end{document}